\documentclass{article}

\usepackage{neurips_2021}

\usepackage[utf8]{inputenc} 
\usepackage[T1]{fontenc}    
\usepackage{hyperref}       
\usepackage{url}            
\usepackage{booktabs}       
\usepackage{amsfonts}       
\usepackage{nicefrac}       
\usepackage{microtype}      
\usepackage{xcolor}         
\usepackage[ruled,vlined]{algorithm2e}
\usepackage{amsmath,amssymb,amsfonts}
\usepackage{algorithmicx}
\usepackage{algpseudocode}
\usepackage{algorithm2e}
\usepackage{graphicx}
\usepackage{textcomp}
\usepackage{xcolor}
\usepackage{tabu}
\usepackage{tabularx}
\usepackage{pdflscape}
\usepackage[font={small,it}]{caption}
\usepackage{wrapfig}
\usepackage{subcaption}
\usepackage{mathtools}
\usepackage{amsthm}

\usepackage{hyperref}
\usepackage[toc,page]{appendix}
\usepackage{multirow}
\newcommand{\rom}[1]{\uppercase\expandafter{\romannumeral #1\relax}}
\SetKwRepeat{Do}{do}{while}
\DeclareMathOperator*{\argmin}{argmin}
\DeclareMathOperator*{\argmax}{argmax}

\title{BO-DBA: Query-Efficient Decision-Based Adversarial Attacks via Bayesian Optimization}

\author{
    Zhuosheng Zhang\\
    Stevens Institute of Technology\\
    zzhang97@stevens.edu\\
    \texttt{zzhang97@stevens.edu} \\
    \and
    Shucheng Yu\\
    Stevens Institute of Technology\\
    \texttt{shucheng.yu@stevens.edu}

}

\begin{document}

\maketitle

\begin{abstract}
Decision-based attacks (DBA), wherein attackers perturb inputs to spoof learning algorithms by observing solely the output labels, are a type of severe adversarial attacks against Deep Neural Networks (DNNs) requiring minimal knowledge of attackers.  State-of-the-art DBA attacks relying on zeroth-order gradient estimation require an excessive number of queries. Recently, Bayesian optimization (BO) has shown promising in reducing the number of queries in score-based attacks (SBA), in which attackers need to observe real-valued probability scores as outputs. However, extending BO to the setting of DBA is nontrivial because in DBA only output labels instead of real-valued scores, as needed by BO, are available to attackers. In this paper, we close this gap by proposing an efficient DBA attack, namely BO-DBA. Different from existing approaches, BO-DBA generates adversarial examples by searching so-called \emph{directions of perturbations}. It then formulates the problem as a BO problem that minimizes the real-valued distortion of perturbations. With the optimized perturbation generation process, BO-DBA converges much faster than the state-of-the-art DBA techniques. Experimental results on pre-trained ImageNet classifiers show that BO-DBA converges within 200 queries while the state-of-the-art DBA techniques need over 15,000 queries to achieve the same level of perturbation distortion. BO-DBA also shows similar attack success rates even as compared to BO-based SBA attacks but with less distortion.
\end{abstract}

\section{Introduction}\label{s:I}

Recent advances in computation and learning have made deep neural networks (DNNs) an important enabler for a plethora of applications. However, DNNs have also shown vulnerabilities to \emph{adversarial examples} - a type of maliciously perturbed examples that are almost identical to original samples in human perception but can cause models to make incorrect decisions\cite{szegedy2013intriguing}. Such vulnerabilities can lead to severe and sometimes fatal consequences in many real-world DNN-powered applications like autonomous vehicles, financial services, and robotics. Therefore, it is critical to understand limitations of current learning algorithms and identify their vulnerabilities, which in turn helps to improve the robustness of learning.

\begin{figure}[h]
\centering
\includegraphics[width=\linewidth]{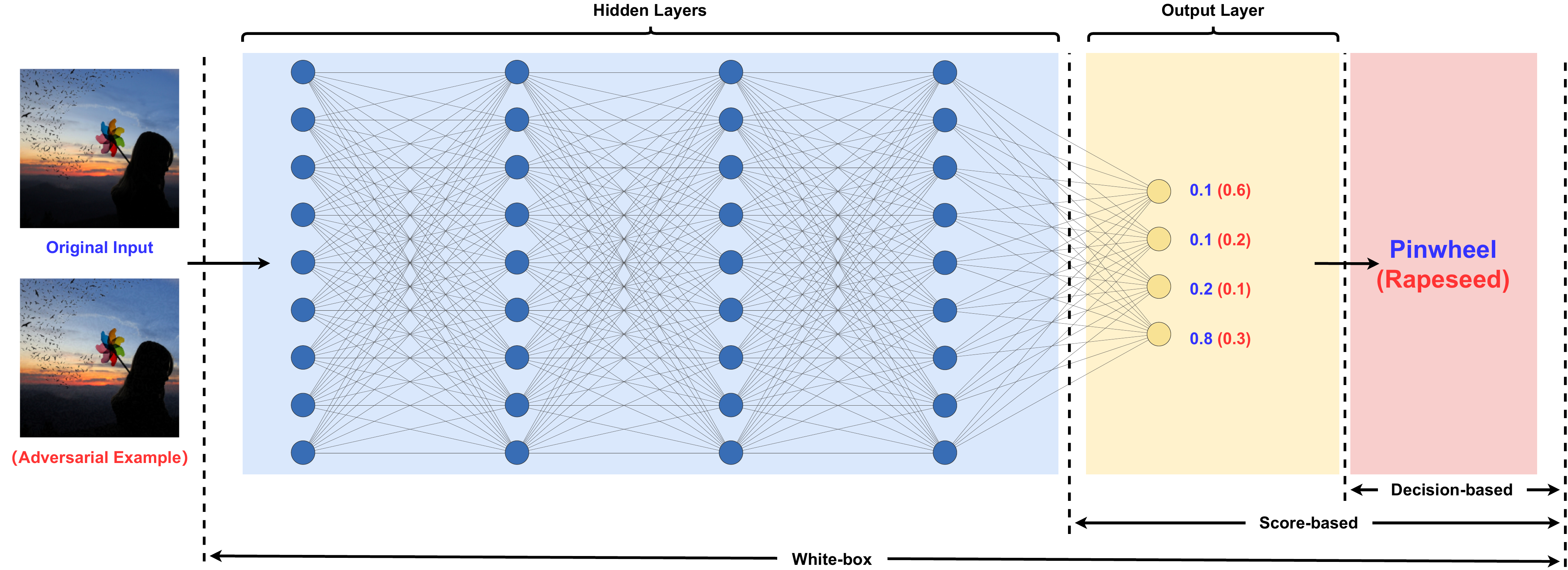}
\caption{An illustration of the three types of adversarial attack. A white-box attack can access to the whole model; a score-based attack can access to the output layer; a decision-based attack can only access the predicted label. With an unnoticeable perturbation, "Pinwheel" is classified as “Rapeseed” }\label{f:ill}
\end{figure}

According to the knowledge of attackers, adversarial attacks can be categorized into two primary types: \emph{black-box attacks} and \emph{white-box attacks}. In the white-box setting\cite{goodfellow2014explaining,madry2017towards}, the attacker requires complete knowledge of the architecture and parameters of the target network. In the black-box setting, the attacker can only access the output(s) of the model.  By the types of outputs, black-box attacks can be further divided into two categories - \emph{score-based attacks} (SBA), in which the outputs are real-valued probability scores, and \emph{decision-based attacks} (DBA), wherein only the output label of a given input is available. Fig. \ref{f:ill} illustrates the accessible information of the target model for each of the three adversarial attacks.


In particular, DBA could lead to severe and ubiquitous threats to practical systems because of the minimal requirement on the knowledge of attackers, and has attracted great interests recently.  To measure the performance of DBA, \emph{query efficiency} and \emph{perturbation quality} are two important matrices that are frequently adopted. Query efficiency measures the number of queries needed to generate adversarial examples per input. Perturbation quality measures the distortion level between adversarial examples and original inputs. Most existing DBA attacks adopt zeroth-order optimization to find adversarial examples \cite{ilyas2018prior,jones1998efficient,chen2017zoo,ilyas2018black,brendel2017decision,chen2020hopskipjumpattack,ilyas2018black,cheng2018query} and require tens of thousands queries before a perturbed example is generated.  Recently, Bayesian Optimization (BO) has been adopted in several proposals \cite{co2019procedural,ru2019bayesopt,shukla2019black} to optimize adversarial attacks and can reduce the number of queries to the order of hundreds. However, these works are orientated to SBA because of their reliance on real-valued objective functions such as probability scores to guide the Bayesian optimizer. Extending BO to the setting of DBA is nontrivial because in DBA only output labels instead of real-valued scores are available to attackers.


In this paper, we close this gap by proposing an efficient DBA attack, namely BO-DBA. Different from existing approaches, BO-DBA generates adversarial examples by searching so-called \emph{directions of perturbations}. It then formulates the problem as a BO problem that minimizes the real-valued distortion of perturbations. With the optimized perturbation generation process, BO-DBA converges much faster than the state-of-the-art DBA techniques \cite{chen2020hopskipjumpattack,cheng2018query}. Experimental results on pre-trained ImageNet classifiers \cite{deng2009imagenet} show that BO-DBA converges within 200 queries while existing DBA techniques \cite{chen2020hopskipjumpattack,cheng2018query} need over 15,000 queries to achieve the same level of perturbation distortion. We also compared BO-DBA with existing BO-based SBA designs \cite{co2019procedural,shukla2019black}. Experimental results show that BO-DBA exhibits similar attack success rates as compared to these BO-based SBA designs but with less distortion. To our best knowledge, BO-DBA is the first BO-based DBA design which exhibits much faster perturbation generation in terms of query numbers than existing DBA techniques.




This rest of the paper is organized as follows: Section \ref{s:RW} reviews adversarial attacks and technical preliminaries. Section \ref{s:TI} describes our technical intuition and Section \ref{s:detail} elaborates the design of BO-DBA. Experimental results are provided in Section \ref{s:E}. We conclude the paper in Section \ref{s:C}.

\section{Related work}\label{s:RW}
\subsection{Black-Box Adversial attacks}
Black-box attacks are one type of adversarial attacks against learning systems where the attacker has no knowledge about the model and can only observe inputs and their corresponding output by querying the model. According to the types of model outputs, black-box attacks can be classified into \textbf{decision-based attacks} (DBA) and \textbf{score-based attacks} (SBA). SBA attackers can access the real-valued probability or the score of each output class while DBA attackers only can access the output labels which may not necessarily be real-valued.

\textbf{DBA} attacks are detrimental to learning systems because the minimal requirements on the knowledge of attackers. There have been several DBA techniques in the literature. In Boundary Attack \cite{brendel2017decision}, a perturbed example is generated starting with a large perturbation sampled from a proposed distribution. It then iteratively reduces the distance of the perturbed example to the original input through a random walk along the decision boundary. With this technique, the Boundary Attack achieves a comparable performance as the state-of-the-art white-box attacks \cite{moosavi2016deepfool,kurakin2016adversarial,gu2014towards} after millions of model queries. Opt-Attack\cite{cheng2018query} improves the efficiency of DBA through exploitation of discarded information with the Zeroth-order gradient estimation. HSJA\cite{chen2020hopskipjumpattack} furthermore improves clock-time efficiency by integrating the majority of model queries into mini-batches. However, one outstanding limitation with existing DBA techniques is the high query complexity. To spoof an undefended ImageNet model, these methods usually need tens of thousands to millions of model queries.

\textbf{SBA} attacks assume the attacker can access the real-valued confidence scores such as class probabilities. In parallel to Boundary attack, GenAttack \cite{alzantot2019genattack} uses genetic algorithms to iteratively evolve a population of candidate adversarial examples. The number of queries needed in this technique is only $1.3\%$ of that in the Boundary Attack. Another design ZOO \cite{chen2017zoo} also adopts the zeroth-order gradient estimation to optimize the confidence score of perturbed inputs. In addition to these two methods, there have been several other techniques using various ways to improve the query efficiency. For example, \cite{tu2019autozoom} performs dimension reduction via Autoencoder; \cite{ilyas2018black} adopts efficient gradient estimation methods together with projected gradient descent to optimize the perturbed example generation. One breakthrough of SBA design was made by the BayesOpt Attack \cite{ru2019bayesopt} and the ``Black-Box Attack" \cite{shukla2019black} made, which only need less than a thousand queries to fool the ImageNet classifier. Both attacks perform Bayesian optimization on some low dimensional inputs and then up-sample them to perturbations using traditional image up-sampling algorithm or deep generative models. The Procedural Attack\cite{co2019procedural} further improves the attack success rate by designing a perturbation generator that can produce natural details using procedural noises \cite{lagae2010survey}. 

\subsection{Bayesian Optimization}
Bayesian optimization (BO) is a sequential optimization method particularly suitable for problems with low dimension and expensive query budgets \cite{mockus2012bayesian} such as black-box optimization. It contains two main components - a \emph{probabilistic surrogate model}, usually a Gaussian Process (GP), for approximating the objective function, and an \emph{acquisition function} that assign a value to each query that describes how optimal the query is.

\textbf{Gaussian Process} is a statistic surrogate that induces a posterior distribution over the objective functions \cite{rasmussen2003gaussian}. In particular, a Guassian Process $\mathcal{GP}(\mu_0,\Sigma_0)$ can be described by a prior mean function $\mu_0$ and positive-definite kernel or covariance function $\Sigma_0$. In this paper, we adapt the Matern $5/2$ Kernel \cite{shahriari2015taking} as the covariance function, which is defined as:
\begin{align*}
    \Sigma(x,x')=(1+\frac{\sqrt{5}r}{l}+\frac{5r^2}{3l^2})exp(-\frac{\sqrt{5}r}{l})
\end{align*}
where $r = x-x'$ and $l$ is the length-scale parameter \cite{snoek2012practical}.

\textbf{Acquisition Function} in Bayesian optimization is a function that evaluates the utility of model querying at each point, given the surrogate model, to find the optimal candidate query point for the next iteration \cite{brochu2010tutorial}. \emph{Expected Improvement} (EI) and \emph{Upper Confidence Bound} (UCB) are the two most popular acquisition functions that have been shown effective in real black-box optimization problems \cite{shahriari2015taking}. In black-box adversarial attacks, most studies \cite{shukla2019black,co2019procedural} adopted EI as the acquisition function because of its better convergence performance \cite{shahriari2015taking,snoek2012practical}. In this paper, we also use EI as the acquisition function which is defined as:
\begin{align*}
    EI_n(x)=\mathbb{E}_n[max(h(x)-h^*_n,0)]
\end{align*}
where $h$ is the objective function and $h^*_n$ is the best observed value so far. $\mathbb{E}_n[\cdot]=\mathbb{E}_n[\cdot|D_{1:n-1}]$ denotes the expectation taken over the posterior distribution given evaluations of $h$ at $x_1,\cdots,x_{n-1}$.
\section{Technical Intuition}\label{s:TI}
Although BO-based SBA has shows promising performance, in a real-world system the real-valued confidence scores or probabilities are not necessarily observable. To understand the full potential of DBA, in this work we design a framework that introduces a new perturbation generator and integrate the Bayesian optimization into the decision-based setting. The key design of our framework is that, instead of relying on the confidence score, we optimize the distance between the perturbed example to the original input using the new perturbation generator, and formulate perturbation optimization problem as a BO problem. As compared to optimizing confidence scores in SBA, the new design is expected to result in an improved perturbation quality because the latter directly optimizes the perturbation. Moreover, due to the optimized perturbation generation process, model queries are made more efficient and less queries will be needed for a given distortion budget.

\begin{figure}[htbp]
\centerline{\includegraphics[width=0.5\linewidth]{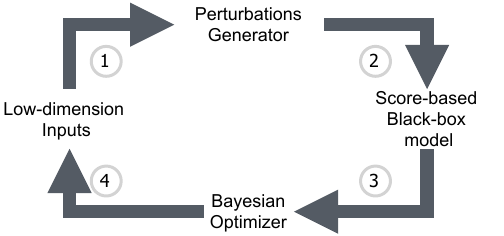}}
\caption{Workflow of BO-based SBA}\label{f:flow}
\end{figure}

\subsection{Difficulties of Applying BO in DBA}
Figure (\ref{f:flow}) shows the general workflow of BO-based attacks in the score-based setting. It contains two critical components - the \textbf{perturbation generator} and the \textbf{Bayesian optimizer}. Perturbation generator is a function $S: \mathbb{R}^{d'}\to \mathbb{R}^d$ that takes low dimensional inputs $\delta' \in \mathbb{R}^{d'}$ and outputs an image-size perturbation $\delta \in \mathbb{R}^d$ (we will take images as examples in the rest of this paper). The score-based model $F$ takes images as inputs and outputs a $K-$dimensional vector which represents confidence scores over $K-$classes. Then the objective function can be defined as following:
\begin{equation}\label{eq:obj1}
  h(\delta') =
    \begin{cases}
      \max\limits_{i\neq y}[F_i(x+S(\delta')] - F_y(x+S(\delta') & \text{(Untargeted)}\\
      F_k(x+S(\delta') - \max\limits_{i\neq k}[F_i(x+S(\delta')] & \text{(Targeted)}\\
    \end{cases}       
\end{equation}

where $x \in \mathbb{R}^d$ is the targeted data sample and $y \in \{1...K\}$ is its true label. $k \in \{1...K\}$ represents the target label. Notes that only adversarial perturbation outputs a positive objective value. Then we can define the following optimization problem to find adversarial perturbation:
\begin{equation}\label{eq:opt1}
    \max_{\delta'}\, h\{x+S(\delta')\}\; \text{ s.t. }\; |S(\delta')|_p \leq \eta
\end{equation}
where $|\cdot|_p$ indicates the desired distance norm that depends on the scenario. The optimization procedural can be summarized in as following: given input $\delta'$, the perturbation generator will first generate noise $S(\delta')$ with the distance constraint; then the attacker directly observes the output of the objective function via the score-based model. Given existing observations, BO will update the posterior probability distribution on the objective function $h$ by GP. Then it will find the optimal low dimensional input $\delta'$ for the next query via optimizing the acquisition function over the current posterior distribution of GP. This procedure will repeat until a valid adversarial example is found or it exceeds the query budget.

In essence, the perturbation generator reduces the searching space from millions of dimensions to less than a hundred dimensions, and the Bayesian optimizer searches for the optimal point in the reduced dimension. However, a DBA attacker may not be able to query the probability score from the model to access $h(x)$. Instead, the attacker may only be able to access the Boolean-valued function $h_b(x)=sign(h(x))$. GP is unable to model an accurate distribution because $h_b(x)$ only returns -1 before the adversarial examples are returned. Moreover, the objective function $h(\cdot)$ in (\ref{eq:opt1}) does not directly relate to the quality of perturbation which is determined by the constraint in (\ref{eq:opt1}). In practice, how to choose the constraint will become a tricky problem.

\subsection{Problem Formulation}
Considering the above difficulties, we propose a new DBA algorithm that is compatible with the Bayesian optimization. Specifically, for a given input $x_0$, perturbation generator $S$ and Boolean-valued function $h_b(x)=sign(h(x))$ defined in decision-based setting, we formulate our objective function $g'(\delta')$ as:
\begin{align}\label{eq:dist2}
    g'(\delta') = \argmin_{\Delta>0} \Big(h_b(x_0+\Delta\theta) = 1\Big) \text{ s.t. } \theta=\frac{S(\delta')}{|S(\delta')|}
\end{align}

In this formulation, $\theta=\frac{S(\delta')}{|S(\delta')|}$ represents the search direction and $g'(\delta')$ is the distance from $x_0$ to the nearest adversarial example along the direction $\theta$. Note that the search direction $\theta$ is the normalized perturbation generated by $S$ and the value of $g'$ can be evaluated via multiple decision-based queries which we will discuss in section \ref{s:detail}. 

Instead of finding the nearest adversarial example directly, we define the following optimization problem to search the optimal direction $\theta$ to get the minimal distortion:
\begin{equation}\label{eq:opt3}
    \min_{\delta'} g'(\delta')
\end{equation}

With (\ref{eq:opt3}), we are able to find the adversarial example efficiently in decision-based setting. Note that, this algorithm can also be applied to the score-based setting to improve the perturbation quality.

\section{Decision-based Bayesian Optimization Attack}\label{s:detail}
In this section, we describe an optimization framework for finding adversarial instances for a classification model $F$ in detail. First we discuss how to compute $g'(\delta')$ up to certain accuracy using the Boolean-valued function $h_b$. Then we will solve the optimization problem via Bayesian Optimization and present our full algorithm.

\begin{algorithm}[!htb]
    \caption{Distance Evaluation Algorithm}\label{alg1}
    \SetAlgoLined
    \SetKwInOut{Input}{input}
    \SetKwInOut{Output}{output}
    \Input{Boolean-valued query function $h_b$ of target model, original image $x_0$, low dimensional input $\delta'$, increase step size $\eta$, stopping tolerance $\epsilon$, maximum distance $\Delta_{max}$}
    \Output{$g'(\delta')$}
    $\theta\gets\frac{S(\delta')}{S(\delta')}$\tcp*{Compute the searching direction}
    \tcp{Fine-grained search}
    \eIf{$h_b(x_0+\eta\theta)=-1$}{
        $v_{low}\gets x_0+\eta\theta,\,v_{high}\gets x_0+2\eta\theta$\;
        \While{$h_b(v_{high})=-1$}
        {
            $v_{low}\gets v_{high},v_{high}\gets v_{high}+\eta\theta$\;
            \If{$|v_{low}|\geq\Delta_{max}$}
            {
                \Return $g'(\delta')=\Delta_{max}$\;
            }
        }
    }
    {
        $v_{low}\gets 0,\,v_{high}\gets x_0+\eta\theta$\;
    }
    \tcp{Binary search between $[v_{low},v_{high}]$}
    \While{$|v_{high}-v_{low}>\epsilon|$}
    {
        $v_{mid}\gets(v_{high}+v_{low})/2$\;
        \eIf{$h_b(v_{mid})=-1$}
        {
            $v_{high}\gets v_{mid}$\;
        }
        {
            $v_{low}\gets v_{mid}$\;
        }
    }
    \Return $g'(\delta')=|v_{high}|$\;
    
\end{algorithm}

\subsection{Distance Evaluation Algorithm}
Algorithm \ref{alg1} elaborates how to evaluate $g'(\delta')$ via queries on Boolean-value function $h_b$:

First, the attacker computes the search direction locally $\theta = \frac{S(\delta')}{|S(\delta')|}$. For a given low dimensional input $\delta'$, attacker first generates an image-size perturbation $S(\delta')$ via the perturbation generator $S$. Then normalize it into a unit vector to represent the search direction. It is easy to notice that for any given input $\delta'$, there is always a search direction $\theta$ that can be computed.

To evaluate the distance from input $x_0$ to the decision boundary along the direction $\theta$, the attacker performs a fine-grain search and then a binary search. For simplicity, we assume the $l_2$ distance here, but the same procedure can be applied to other distance measurements as well as long as vector operations are well defined in their respective spaces. In the fine-grained search phase, we cumulatively increase the searching distance to query the points $\{x_0+\eta\theta,x_0+2\eta\theta,\dots\}$ one by one until $h_b(x_0+i\eta\theta)=1$. Then we conduct a binary search between the interval $[x_0+(i-1)\eta\theta,x_0+i\eta\theta]$, within which the classification boundary is located. Note that, in practice the fine-grained search may exceed the numerical bounds defined by the image (or other type of samples). We can simply assign a maximum distance (e.g., the distance between all-black image and all-white image) for this searching direction. Unlike the gradient-based method that needs an accurate result to evaluate the gradient, Bayesian Optimization only needs statistical knowledge about each direction.

\subsection{Bayesian Optimization}
With the real-valued function $g'(\delta')$ available, the optimization problem (\ref{eq:opt3}) is solvable using the Bayesian Optimization. Adopting the similar logic in Figure \ref{f:flow}, the attacker will query the boundary distance on the searching direction generated by the low dimensional input. The optimizer models the distribution of distances over the input space and acquires the next most possible optimal input for querying until an adversarial example near enough is found.
\begin{algorithm}[!htb]
    \caption{Bayesian Optimization for DBA}\label{alg2}
    \SetAlgoLined
    \SetKwInOut{Input}{input}
    \SetKwInOut{Output}{output}
    \Input{Targeted input $x_0$, Guassian process model GP, Acquisition function $\mathcal{A}$, Initialization sample size $T_0$, Maximum sample size $T$, Distance evaluation function $g'(\cdot)$, stopping tolerance  $\epsilon$, $D=\varnothing$.}
    \Output{Adversarial Examples $x'$}
    \tcp{Intialization}
    \For {$t=0,1,2...,T_0-1$}
    {
        Generate input $\delta'_t$ randomly\;
        $D\gets D\cup (\delta'_t,g'(\delta'_t))$\;
    }
    Update the GP on $D$\;
    \tcp{Optimization via GP and Acquisition function}
    \While{$t<T$}
    {
        $t\gets t+1$\;
        $\delta'_t\gets \argmax_{\delta'}\mathcal{A}(\delta',D)$\;
        \eIf{$|g'(\delta'_t)|>\epsilon$}
        {
            $D\gets D\cup (\delta'_t,g'(\delta'_t))$\ and update the GP\;    
        }
        {
            $\theta=\frac{S(\delta'_t)}{|S(\delta'_t)|}$\;
            \Return $x_0+g'(\delta'_t)\theta$\;
        }

    }
    \tcp{Return nearest adversarial example}
    $\theta=\frac{S(\delta'_{*})}{|S(\delta'_{*})|}\, |\,  (\delta'_{*},g(\delta'_{*})) \in D  \textbf{ such that } g(\delta'_{*}) \leq g(\delta')\quad \forall (\delta',g'(\delta')) \in D$\;
    \Return $x_0+g'(\delta'_{*})\theta$\;
    
\end{algorithm}

The detailed procedural is presented in Algorithm \ref{alg2}. At beginning, we sample $T_0$ random low dimensional inputs $\delta'$ from the input space and evaluate the distance $g'(\delta')$ using Algorithm \ref{alg1}. Then we iteratively update the posterior distribution of the GP using available data $D$ and query new $\delta'$ obtained by maximizing the acquisition function over the current posterior distribution of GP until a valid adversarial example within the desired distortion is found or the maximum number of iteration is reached. Note that the query budget shall be larger than the number of iterations because we need multiple queries to evaluate the distance in Algorithm \ref{alg1}. The alternative stop condition of the optimization procedure is to set a maximum acceptable query budget. 

Compare to the score-based methods, our optimization framework focus on optimizing the boundary distance which can be evaluated via querying the decision-based model solely. Moreover, our algorithm does not rely on the distortion constraints like score-based BO, where the attacker needs to define the required boundary distance beforehand to trade success rate for perturbation quality.

\section{Experiments}\label{s:E}
In this section, we carry out experimental analysis of our work. We compare query efficiency and perturbation quality of our optimization framework with several previously proposed DBA attacks and BO-based SBA attacks. We also evaluate the abilities of bypassing existing defense mechanisms \cite{xu2017feature} for three attack mechanisms. In addition, we examine how different types of perturbation generators affect query efficiency and perturbation quality. All experiments are carried out on a 2080 TI GPU, with code available online.\footnote{\url{https://github.com/zzs1324/BO-DBA.git}}

\subsection{Performance Evaluation}\label{ss:PE}

\textbf{Baselines:} We first compare our design with the state-of-the-art DBA attacks - HSJA \cite{chen2020hopskipjumpattack} and Opt-Attack \cite{cheng2018query}. We also compare our design with the-state-of-art BO-based SBA attacks - Procedural \cite{co2019procedural} and BayesOPT \cite{ru2019bayesopt}.

\textbf{Data and Models:} We use two distinct DCN architectures pre-trained on ImageNet\cite{deng2009imagenet}: ResNet-50\cite{he2016deep} and Inception V3 \cite{szegedy2016rethinking}. ResNet-50 takes input images with dimensions $224\times 224 \times 3$ while Inception V3 take images with dimensions $299\times 299 \times 3$. In addition to defenseless, we also carried out experiments on classifier equipped with runtime adversarial sample detection algorithm\cite{xu2017feature} which detect the abnormal inputs by comparing a DNN model’s prediction on the original input with that on squeezed inputs (reducing the color bit depth of each pixel or spatial smoothing).

\textbf{Metrics:} To measure the efficiency, we use the average $l_2\&l_\infty$ distance between perturbed and original samples over a subset of test images. In addition to $l_2\&l_\infty$ distance, we furthermore measured average perceptual similarity score \textsl{LPIPS}\cite{zhang2018unreasonable} over a subset of test images to evaluate the level of indistinguishability between adversarial examples and original inputs in human preception.

As an alternative metric, we also evaluate the \textsl{attack success rate (ASR)}. An adversarial example is defined a success if the distortion distance does not exceed a given distance threshold. In this paper, we use the distance threshold ($l_\infty\leq \frac{16}{255}$) to define the ASR which is the same constraints used in BO-based SBAs.

\begin{figure*}[htbp]
\centerline{\includegraphics[width=0.9\linewidth]{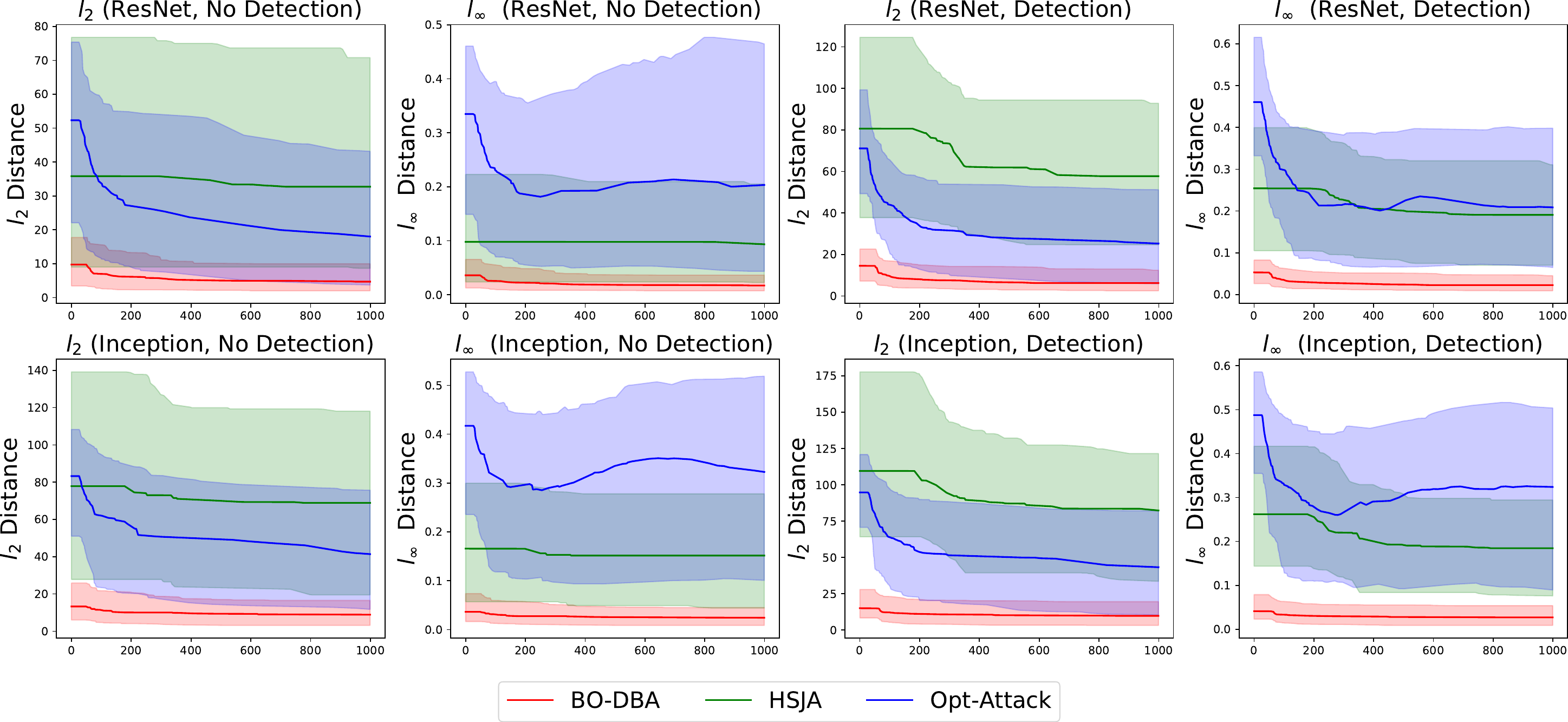}}
\caption{Median distance versus query budgets on ImageNet with ResNet-50 and Inception V3.}\label{f:exp1}
\vspace{-3mm}
\end{figure*}

\textbf{Results:}
Figure \ref{f:exp1} shows the median distance against the query budgets, with the first and third quartiles used as lower and upper error bars. Column $1\&2$ compares the $l_2\&l_\infty$ distortion of our framework against two baseline DBAs on a defenseless classifier. We can see that our algorithm consistently achieves smaller distortion within 1000 queries than baseline methods. Moreover, in ImageNet dataset, our algorithm can converge within 200 query budgets, while the state-of-the-art DBA techniques need over 15,000 queries to achieve the same level of perturbation distortion\cite{chen2020hopskipjumpattack}. In Column $3\&4$, we compare three attacks on the same classifier but equipped with runtime adversarial sample detection algorithm. Although the defense mechanism can increase the necessary distortion level to generate adversarial examples, our algorithm still maintains significantly less distortion than two baseline decision-based attacks within 1000 queries.

In addition to comparing with DBAs, we also compared ASR and perturbation quality between our work and BO-based SBAs in fixed query budgets (300). Table \ref{tab:Compare} shows that BO-DBA can achieve comparable ASR as BO-based SBAs but with less distortion. Moreover, BO-based SBAs will not generate any adversarial examples if distortion constraint is unsatisfied. But BO-DBA will always generate adversarial examples within or exceeding the threshold. For perturbation quality, our methods are minimizing the perturbation distance to generate adversarial examples with less distortion distance, while traditional BO-based SBAs only generate adversarial examples that satisfy the distortion constraints and will not continuously optimize the quality of adversarial examples after finding the adversarial examples. This supports our intuition of reformulating optimization problem in Section \ref{s:TI}. 

Note that, in the case of BO-DBA*(BO-DBA equipped with the same perturbation generator as BayesOPT) vs. BayesOPT, our method has significantly less ASR and relatively higher $l_2\&l_\infty$ distance (but lower LPIPS) than BayesOPT. That's because the input space of those two attacks is 48 dimensions, while, in the case of BO-DBA vs. Procedural, the input space is 3 dimensions. With limited query budgets, BO-DBA* cannot query enough data samples to model the distribution of objective function because BO-DBA spends parts of the queries to evaluate the distortion distance.

\subsection{Effect of Perturbation Generator}
\textbf{Baselines:} We explore the influence of different perturbation generators on attack efficiency and perturbation quality when combining with our framework. In general, we can divide the tested perturbation generators into three types: \textbf{procedural noises generators}\cite{co2019procedural}, \textbf{interpolation-based function}\cite{ru2019bayesopt} and \textbf{clustering-based function }\cite{shukla2019black}.  Procedural noises use the random function to generate an image with complex and intricate details which are widely used in computer graphics\cite{lagae2010survey}. For procedural noises, we consider Perlin and Gabor noise. Interpolation-based functions are widely used in image rescaling and we consider bilinear(BILI) and bicubic(BICU) interpolation. For the clustering-based function, we consider nearest-neighbor (NN) and clustering (CL).

\textbf{Metrics:} In addition to the same evaluation matrix used in Section \ref{ss:PE}, we also measure the inherent properties of perturbation generators like University evasion rate (UAR)\cite{co2019procedural} which refers to perturbation's ability to apply across a dataset or to other models. Given a model $f$, a perturbation $s$, input $x\in X$ and a threshold $\epsilon$, the UAR of s over X is:
\begin{align*}
    \frac{|\{x\in X : \argmax f(x+s)\neq \tau(x)\}|}{|X|}, |s|_{\infty}\leq \epsilon
\end{align*}
Where $\tau(x)$ is the true label of $x$. And we choose the $\epsilon=\frac{16}{255}$.
\begin{table}
	\centering
	\captionsetup{width=\linewidth}
	\caption{Comparison between our method and BO-based SBA. BO-DBA and BO-DBA* are our methods equipped with the same perturbation generator as Procedural and BayesOPT respectively. We compute the distortion of generated adversarial examples up to a certain percentage ($50\%\,75\%\,100\%$), N/A means the percentage of adversarial examples is not enough.}\label{tab:Compare}
	\begin{tabular}{ |c|c|c|c|c|c|c|c|c|c|c| } 
	    \hline
		\multirow{2}{*}{Attack}&\multirow{2}{*}{ASR}&\multicolumn{3}{c|}{LPIPS}&\multicolumn{3}{c|}{$l_2$}&\multicolumn{3}{c|}{$L_\infty$}\\
		\cline{3-11}
		 &  &50&75&100&50&75&100&50&75&100\\
		\hline
        BO-DBA&$84\%$&0.028&0.056&0.11&4.88&6.93&12.23&0.013&0.02&0.04\\
        Procedural&$89\%$&0.205&0.199&N/A&17.21&17.21&N/A&0.06&0.06&N/A\\
        BO-DBA*&$29\%$&0.004&0.102&0.203&16.69&32.97&57.67&0.084&0.18&0.27\\
        BayesOPT&$51\%$&0.016&N/A&N/A&12.63&N/A&N/A&0.06&0.06&N/A\\
		\hline
	\end{tabular}
\vspace{-6mm}
\end{table}

  \begin{minipage}{\textwidth}
  \begin{minipage}[t]{0.4\textwidth}
    \centering

    \includegraphics[width=\linewidth]{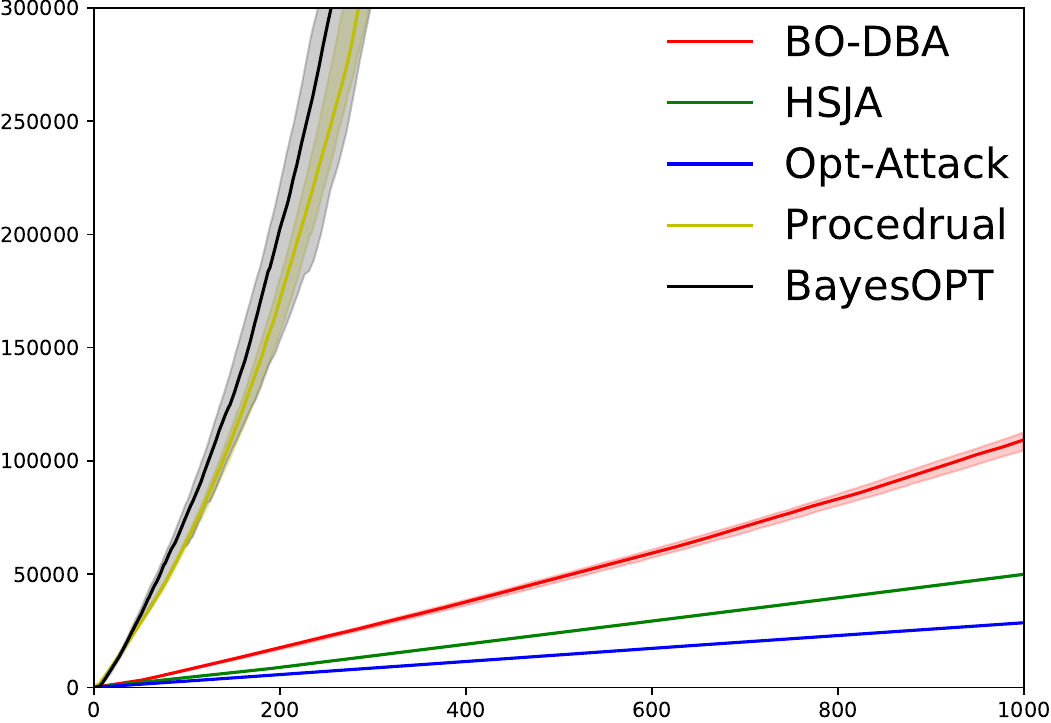}
    \captionof{figure}{Running time (ms) vs Query budgets.}\label{fig:Overhead}
  \end{minipage}
  \hfill
  \begin{minipage}[b]{0.6\textwidth}
    \centering

    	\begin{tabular}{ |c|c|c|c|c|c| } 
		\hline
		Generator & UAR&ASR & LPIPS &$l_2$&$L_\infty$ \\
		\hline
		Perlin & $26.9\%$&$87\%$&0.087&9.28&0.035  \\
		Gabor & $29.4\%$&$77\%$ &0.135&15.70&0.842\\
		BILI & $4.7\%$&$27\%$&0.158&43.44&0.279\\
		BICU & $5.9\%$&$26\%$&0.155&42.14&0.266\\
		CL & $9.0\%$&$67\%$&0.259&18.92&0.862\\
		NN & $11.9\%$&$55\%$&0.115&24.94&0.891\\
		\hline
	\end{tabular}
      \captionof{table}{Perturbation Generators Evaluation}\label{tab:Generator}
    \end{minipage}
  \end{minipage}

\textbf{Results:} Table \ref{tab:Generator} compares the perturbation quality and inherent properties of different perturbation generators. We can see that perturbation generators that belong to the same category have similar inherent properties. For example, procedural noises generators have higher UAR than other generators. In the terms of distortion quality, we found each noise generator has distinct distortion quality in different distortion measurements. Overall, human-designed perturbation generators like procedural noises generators are outperforming the traditional upsampling functions which suggests that designing a good perturbation generator can also improve the attack efficiency and perturbation qualities.  Designing the perturbation generators is beyond the scope of this paper, we will leave it to future works.

\section{Conclusion}\label{s:C}
We introduce a new decision-based attack that leverages Bayesian optimization by formulating the optimization problem to find adversarial perturbations with high query efficiency.  In comparison to existing decision-based attack methods, our BO-DBA converges within 200 queries while the state-of-the-art DBA techniques need over 15,000 queries to achieve the same level of perturbation distortion. BO-DBA also shows similar attack success rates even as compared to BO-based SBA attacks but with less distortion.

One limitation of BO-DBA is that, compared to existing alternatives, attacks equipped with Bayesian Optimization are more computational expensive especially as the query number increases. In Fig. \ref{fig:Overhead}, we compared the running time of five different adversarial attacks. From the plot, we can see gradient-based attacks like HSJA or Opt-Attack have constant computation costs for each iteration. But conventional BO attacks like Procedural or BayesOPT have tremendous computation increase rates. That's because Bayesian Optimization is a non-parametric optimization method with quadratic computation complexity. Although BO-DBA updates the BO model at a lower frequency because part of the model quires is used to evaluate the distortion distance, BO-DBA still has a positive computation increasing rate. Thus, our method is optimal for the scenario where the cost of evaluating the target model or the risk of being detected is much higher than the computational cost of the attack algorithm itself.

On another hand, it's possible to combine our method with gradient-based methods to take advantage of each other. For example, attackers can search possible adversarial examples via our method first then fine-tuning the adversarial example via gradient-based methods.

\bibliographystyle{plain}
\bibliography{citation}

\begin{thebibliography}{10}

\bibitem{alzantot2019genattack}
Moustafa Alzantot, Yash Sharma, Supriyo Chakraborty, Huan Zhang, Cho-Jui Hsieh,
  and Mani~B Srivastava.
\newblock Genattack: Practical black-box attacks with gradient-free
  optimization.
\newblock In {\em Proceedings of the Genetic and Evolutionary Computation
  Conference}, pages 1111--1119, 2019.

\bibitem{brendel2017decision}
Wieland Brendel, Jonas Rauber, and Matthias Bethge.
\newblock Decision-based adversarial attacks: Reliable attacks against
  black-box machine learning models.
\newblock {\em arXiv preprint arXiv:1712.04248}, 2017.

\bibitem{brochu2010tutorial}
Eric Brochu, Vlad~M Cora, and Nando De~Freitas.
\newblock A tutorial on bayesian optimization of expensive cost functions, with
  application to active user modeling and hierarchical reinforcement learning.
\newblock {\em arXiv preprint arXiv:1012.2599}, 2010.

\bibitem{chen2020hopskipjumpattack}
Jianbo Chen, Michael~I Jordan, and Martin~J Wainwright.
\newblock Hopskipjumpattack: A query-efficient decision-based attack.
\newblock In {\em 2020 ieee symposium on security and privacy (sp)}, pages
  1277--1294. IEEE, 2020.

\bibitem{chen2017zoo}
Pin-Yu Chen, Huan Zhang, Yash Sharma, Jinfeng Yi, and Cho-Jui Hsieh.
\newblock Zoo: Zeroth order optimization based black-box attacks to deep neural
  networks without training substitute models.
\newblock In {\em Proceedings of the 10th ACM Workshop on Artificial
  Intelligence and Security}, pages 15--26, 2017.

\bibitem{cheng2018query}
Minhao Cheng, Thong Le, Pin-Yu Chen, Jinfeng Yi, Huan Zhang, and Cho-Jui Hsieh.
\newblock Query-efficient hard-label black-box attack: An optimization-based
  approach.
\newblock {\em arXiv preprint arXiv:1807.04457}, 2018.

\bibitem{co2019procedural}
Kenneth~T Co, Luis Mu{\~n}oz-Gonz{\'a}lez, Sixte de~Maupeou, and Emil~C Lupu.
\newblock Procedural noise adversarial examples for black-box attacks on deep
  convolutional networks.
\newblock In {\em Proceedings of the 2019 ACM SIGSAC Conference on Computer and
  Communications Security}, pages 275--289, 2019.

\bibitem{deng2009imagenet}
Jia Deng, Wei Dong, Richard Socher, Li-Jia Li, Kai Li, and Li~Fei-Fei.
\newblock Imagenet: A large-scale hierarchical image database.
\newblock In {\em 2009 IEEE conference on computer vision and pattern
  recognition}, pages 248--255. Ieee, 2009.

\bibitem{goodfellow2014explaining}
Ian~J Goodfellow, Jonathon Shlens, and Christian Szegedy.
\newblock Explaining and harnessing adversarial examples.
\newblock {\em arXiv preprint arXiv:1412.6572}, 2014.

\bibitem{gu2014towards}
Shixiang Gu and Luca Rigazio.
\newblock Towards deep neural network architectures robust to adversarial
  examples.
\newblock {\em arXiv preprint arXiv:1412.5068}, 2014.

\bibitem{he2016deep}
Kaiming He, Xiangyu Zhang, Shaoqing Ren, and Jian Sun.
\newblock Deep residual learning for image recognition.
\newblock In {\em Proceedings of the IEEE conference on computer vision and
  pattern recognition}, pages 770--778, 2016.

\bibitem{ilyas2018black}
Andrew Ilyas, Logan Engstrom, Anish Athalye, and Jessy Lin.
\newblock Black-box adversarial attacks with limited queries and information.
\newblock In {\em International Conference on Machine Learning}, pages
  2137--2146. PMLR, 2018.

\bibitem{ilyas2018prior}
Andrew Ilyas, Logan Engstrom, and Aleksander Madry.
\newblock Prior convictions: Black-box adversarial attacks with bandits and
  priors.
\newblock {\em arXiv preprint arXiv:1807.07978}, 2018.

\bibitem{jones1998efficient}
Donald~R Jones, Matthias Schonlau, and William~J Welch.
\newblock Efficient global optimization of expensive black-box functions.
\newblock {\em Journal of Global optimization}, 13(4):455--492, 1998.

\bibitem{kurakin2016adversarial}
Alexey Kurakin, Ian Goodfellow, Samy Bengio, et~al.
\newblock Adversarial examples in the physical world, 2016.

\bibitem{lagae2010survey}
Ares Lagae, Sylvain Lefebvre, Rob Cook, Tony DeRose, George Drettakis, David~S
  Ebert, John~P Lewis, Ken Perlin, and Matthias Zwicker.
\newblock A survey of procedural noise functions.
\newblock In {\em Computer Graphics Forum}, volume~29, pages 2579--2600. Wiley
  Online Library, 2010.

\bibitem{madry2017towards}
Aleksander Madry, Aleksandar Makelov, Ludwig Schmidt, Dimitris Tsipras, and
  Adrian Vladu.
\newblock Towards deep learning models resistant to adversarial attacks.
\newblock {\em arXiv preprint arXiv:1706.06083}, 2017.

\bibitem{mockus2012bayesian}
Jonas Mockus.
\newblock {\em Bayesian approach to global optimization: theory and
  applications}, volume~37.
\newblock Springer Science \& Business Media, 2012.

\bibitem{moosavi2016deepfool}
Seyed-Mohsen Moosavi-Dezfooli, Alhussein Fawzi, and Pascal Frossard.
\newblock Deepfool: a simple and accurate method to fool deep neural networks.
\newblock In {\em Proceedings of the IEEE conference on computer vision and
  pattern recognition}, pages 2574--2582, 2016.

\bibitem{rasmussen2003gaussian}
Carl~Edward Rasmussen.
\newblock Gaussian processes in machine learning.
\newblock In {\em Summer School on Machine Learning}, pages 63--71. Springer,
  2003.

\bibitem{ru2019bayesopt}
Binxin Ru, Adam Cobb, Arno Blaas, and Yarin Gal.
\newblock Bayesopt adversarial attack.
\newblock In {\em International Conference on Learning Representations}, 2019.

\bibitem{shahriari2015taking}
Bobak Shahriari, Kevin Swersky, Ziyu Wang, Ryan~P Adams, and Nando De~Freitas.
\newblock Taking the human out of the loop: A review of bayesian optimization.
\newblock {\em Proceedings of the IEEE}, 104(1):148--175, 2015.

\bibitem{shukla2019black}
Satya~Narayan Shukla, Anit~Kumar Sahu, Devin Willmott, and J~Zico Kolter.
\newblock Black-box adversarial attacks with bayesian optimization.
\newblock {\em arXiv preprint arXiv:1909.13857}, 2019.

\bibitem{snoek2012practical}
Jasper Snoek, Hugo Larochelle, and Ryan~P Adams.
\newblock Practical bayesian optimization of machine learning algorithms.
\newblock In {\em Advances in neural information processing systems}, pages
  2951--2959, 2012.

\bibitem{szegedy2016rethinking}
Christian Szegedy, Vincent Vanhoucke, Sergey Ioffe, Jon Shlens, and Zbigniew
  Wojna.
\newblock Rethinking the inception architecture for computer vision.
\newblock In {\em Proceedings of the IEEE conference on computer vision and
  pattern recognition}, pages 2818--2826, 2016.

\bibitem{szegedy2013intriguing}
Christian Szegedy, Wojciech Zaremba, Ilya Sutskever, Joan Bruna, Dumitru Erhan,
  Ian Goodfellow, and Rob Fergus.
\newblock Intriguing properties of neural networks.
\newblock {\em arXiv preprint arXiv:1312.6199}, 2013.

\bibitem{tu2019autozoom}
Chun-Chen Tu, Paishun Ting, Pin-Yu Chen, Sijia Liu, Huan Zhang, Jinfeng Yi,
  Cho-Jui Hsieh, and Shin-Ming Cheng.
\newblock Autozoom: Autoencoder-based zeroth order optimization method for
  attacking black-box neural networks.
\newblock In {\em Proceedings of the AAAI Conference on Artificial
  Intelligence}, volume~33, pages 742--749, 2019.

\bibitem{xu2017feature}
Weilin Xu, David Evans, and Yanjun Qi.
\newblock Feature squeezing: Detecting adversarial examples in deep neural
  networks.
\newblock {\em arXiv preprint arXiv:1704.01155}, 2017.

\bibitem{zhang2018unreasonable}
Richard Zhang, Phillip Isola, Alexei~A Efros, Eli Shechtman, and Oliver Wang.
\newblock The unreasonable effectiveness of deep features as a perceptual
  metric.
\newblock In {\em Proceedings of the IEEE conference on computer vision and
  pattern recognition}, pages 586--595, 2018.

\end{thebibliography}


\appendix

\section{Adversarial Examples}
Fig. \ref{fig:Apdx1} contains some samples of adversarial examples from BO-DBA attacks and other attacks we compared in Section \ref{s:E} against image classification tasks. Fig. \ref{fig:Apdx2} contains some samples of adversarial examples from BO-DBA attacks with different perturbation generators we evaluated in Section \ref{s:E}.

\begin{figure}[htbp]
    \centering
    \begin{subfigure}{0.9\textwidth}
     \includegraphics[width=\linewidth]{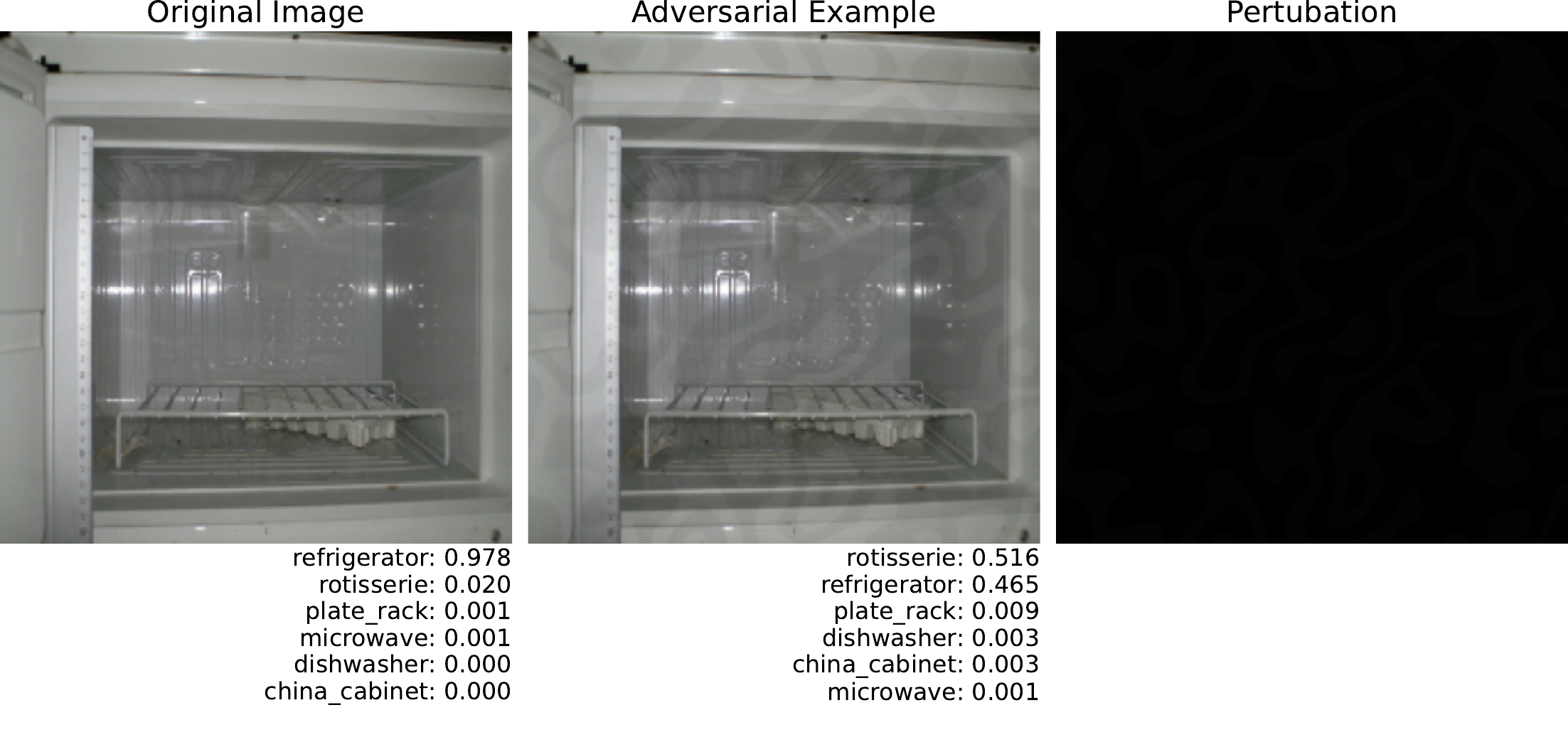}
     \caption{BO-DBA}
    \end{subfigure}
\end{figure}
\begin{figure}\ContinuedFloat
     \centering
    \begin{subfigure}{0.9\textwidth}
     \includegraphics[width=\linewidth]{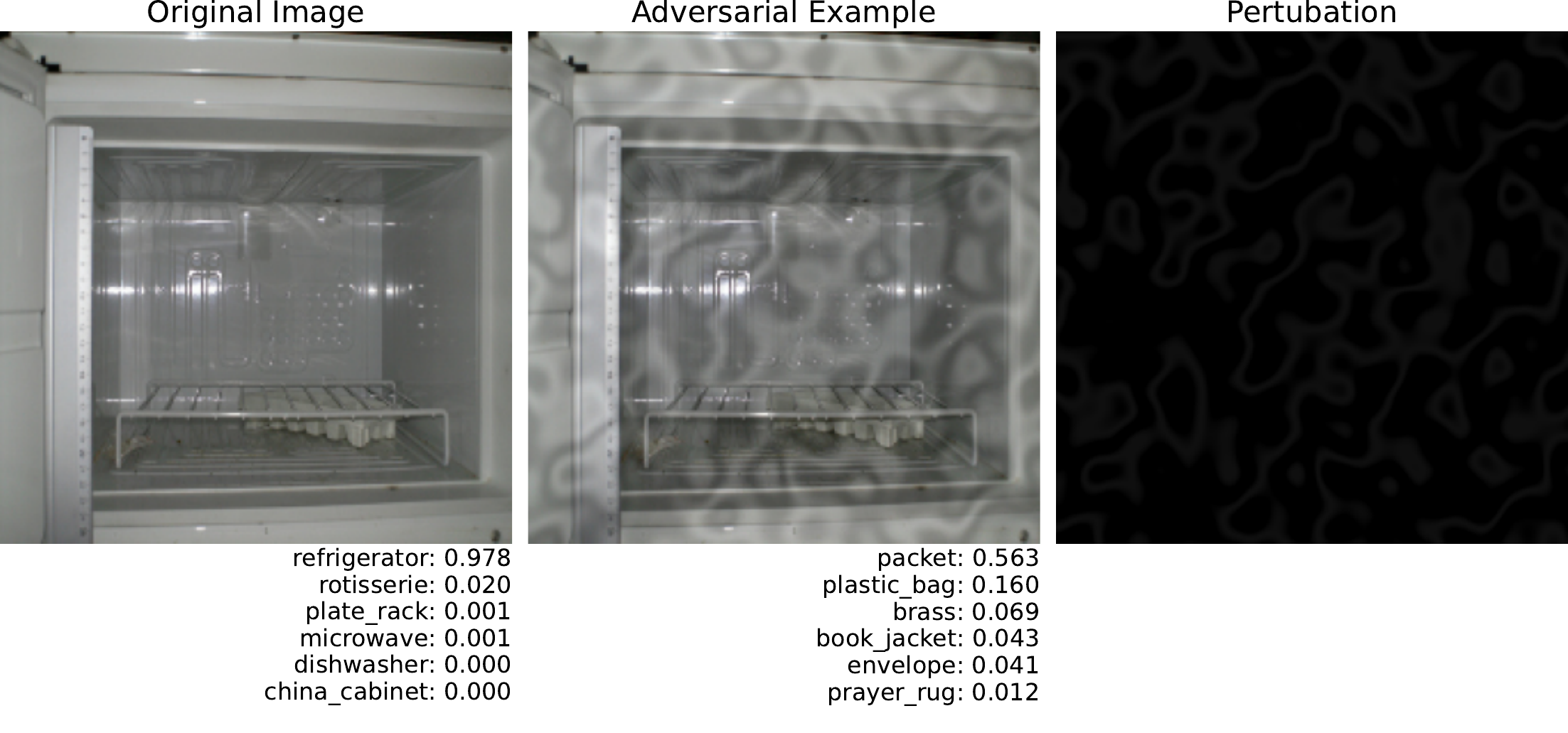}
     \caption{Procedural}
    \end{subfigure}
\end{figure}
\begin{figure}\ContinuedFloat
     \centering
    \begin{subfigure}{0.9\textwidth}
     \includegraphics[width=\linewidth]{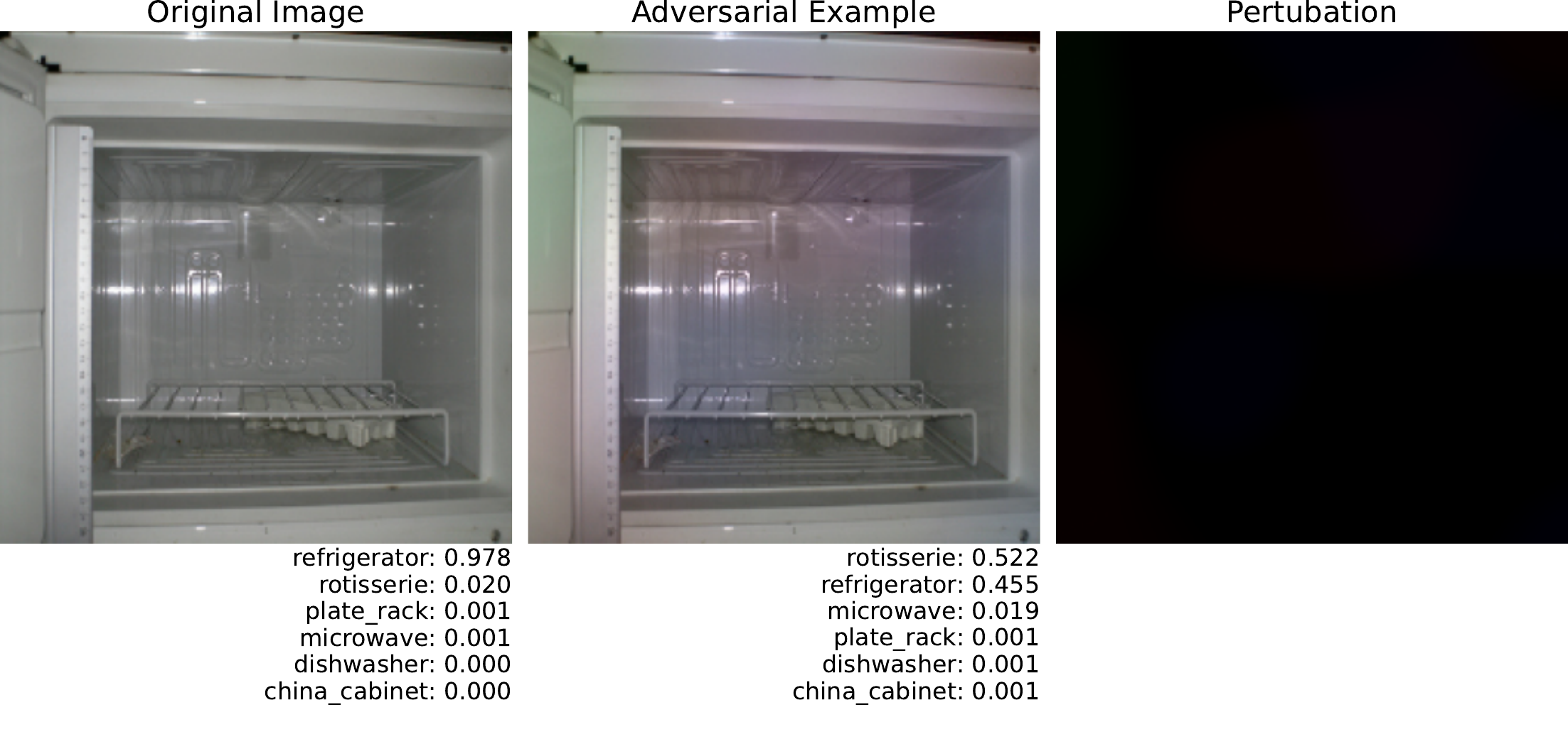}
     \caption{BO-DBA with BICU Generator}
    \end{subfigure}
\end{figure}
\begin{figure}\ContinuedFloat
    \centering
    \begin{subfigure}{0.9\textwidth}
     \includegraphics[width=\linewidth]{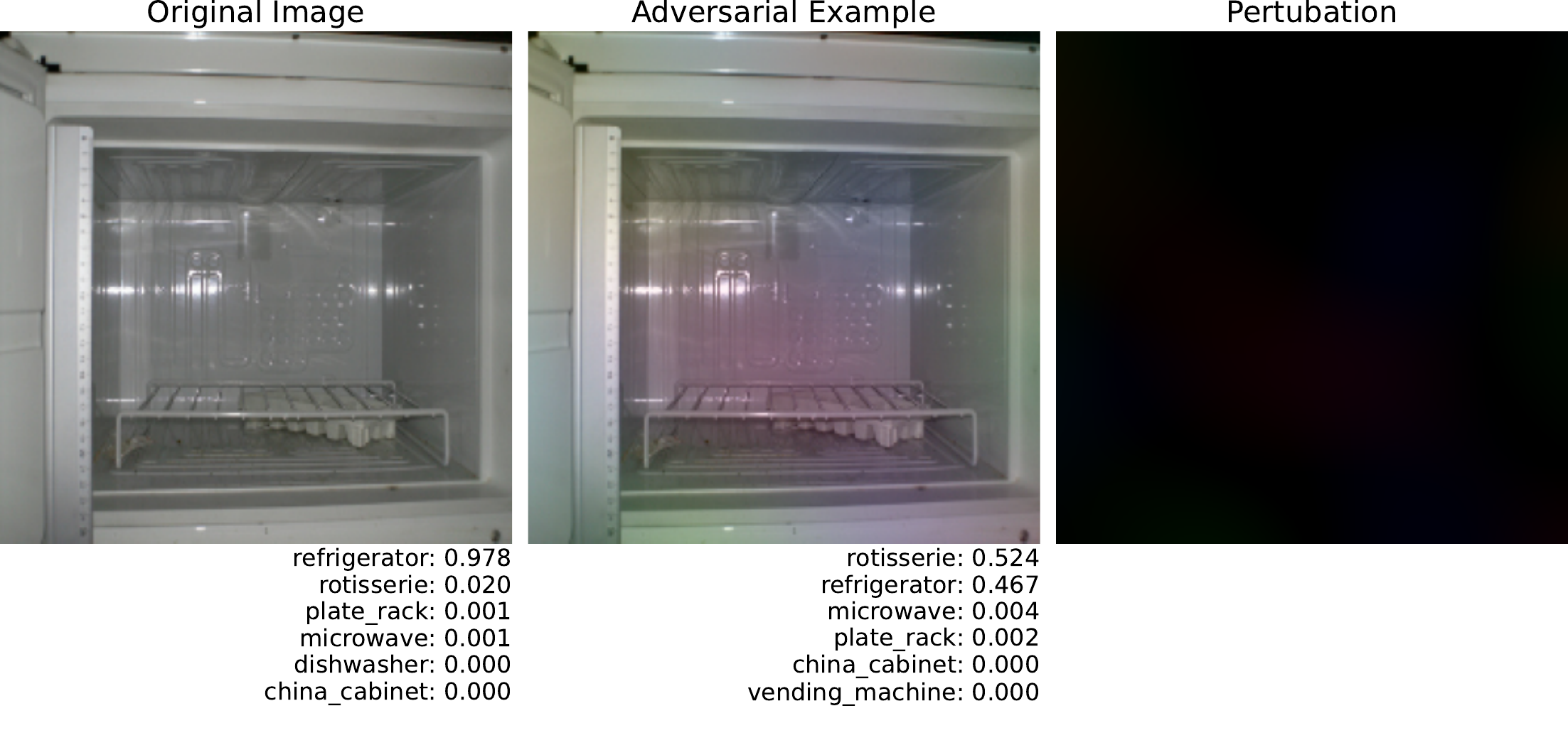}
     \caption{BayesOPT}
    \end{subfigure}
\end{figure}
\begin{figure}\ContinuedFloat
    \centering
    \begin{subfigure}{0.9\textwidth}
     \includegraphics[width=\linewidth]{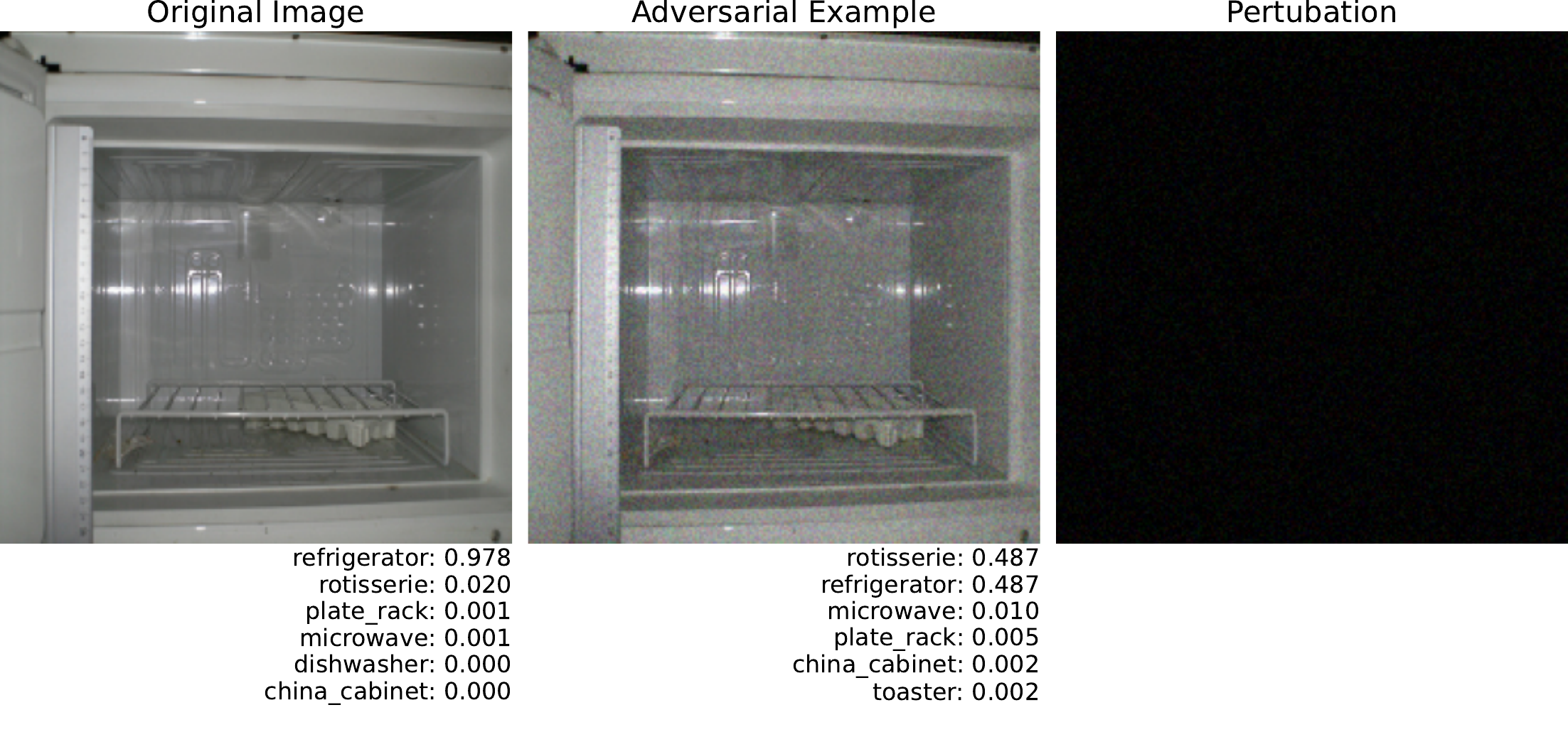}
     \caption{HJSA}
    \end{subfigure}
\end{figure}
\begin{figure}\ContinuedFloat
    \centering
    \begin{subfigure}{0.9\textwidth}
     \includegraphics[width=\linewidth]{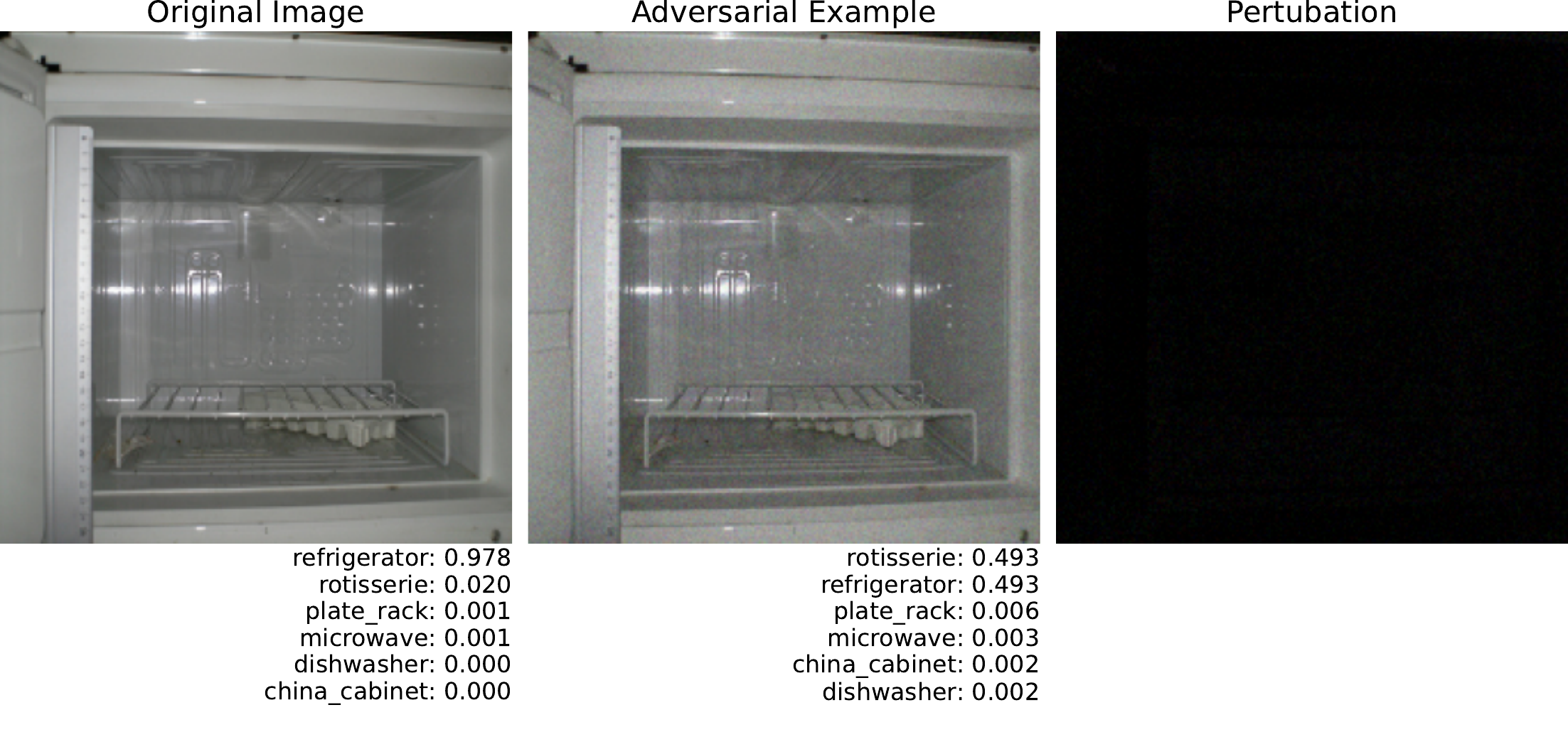}
     \caption{Opt-Attack}
    \end{subfigure}
    \caption{Adversarial examples generated by different algorithms}\label{fig:Apdx1}
\end{figure}

\begin{figure}[htbp]
    \centering
    \begin{subfigure}{0.9\textwidth}
     \includegraphics[width=\linewidth]{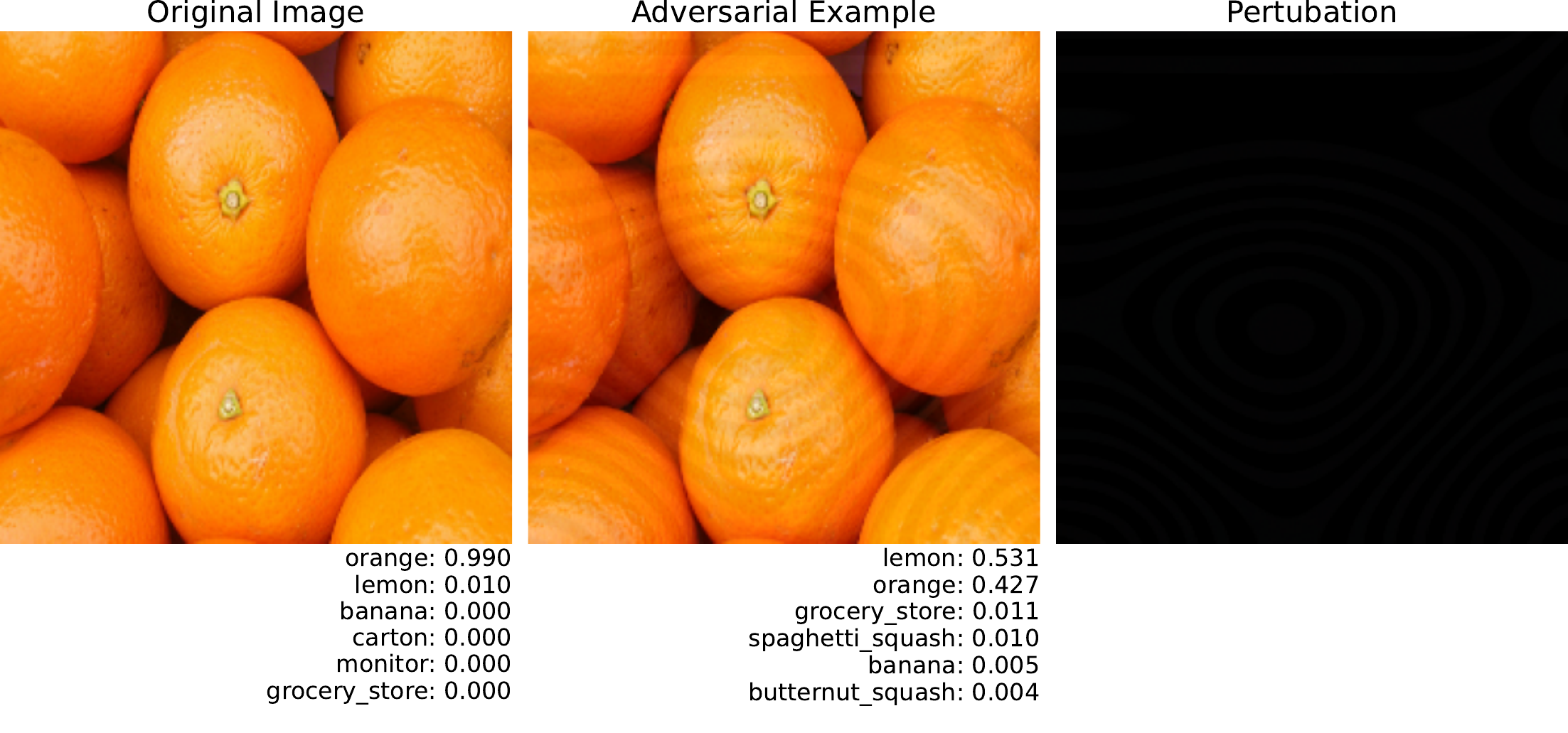}
     \caption{BO-DBA with Perlin generator}
    \end{subfigure}
\end{figure}
\begin{figure}\ContinuedFloat
     \centering
    \begin{subfigure}{0.9\textwidth}
     \includegraphics[width=\linewidth]{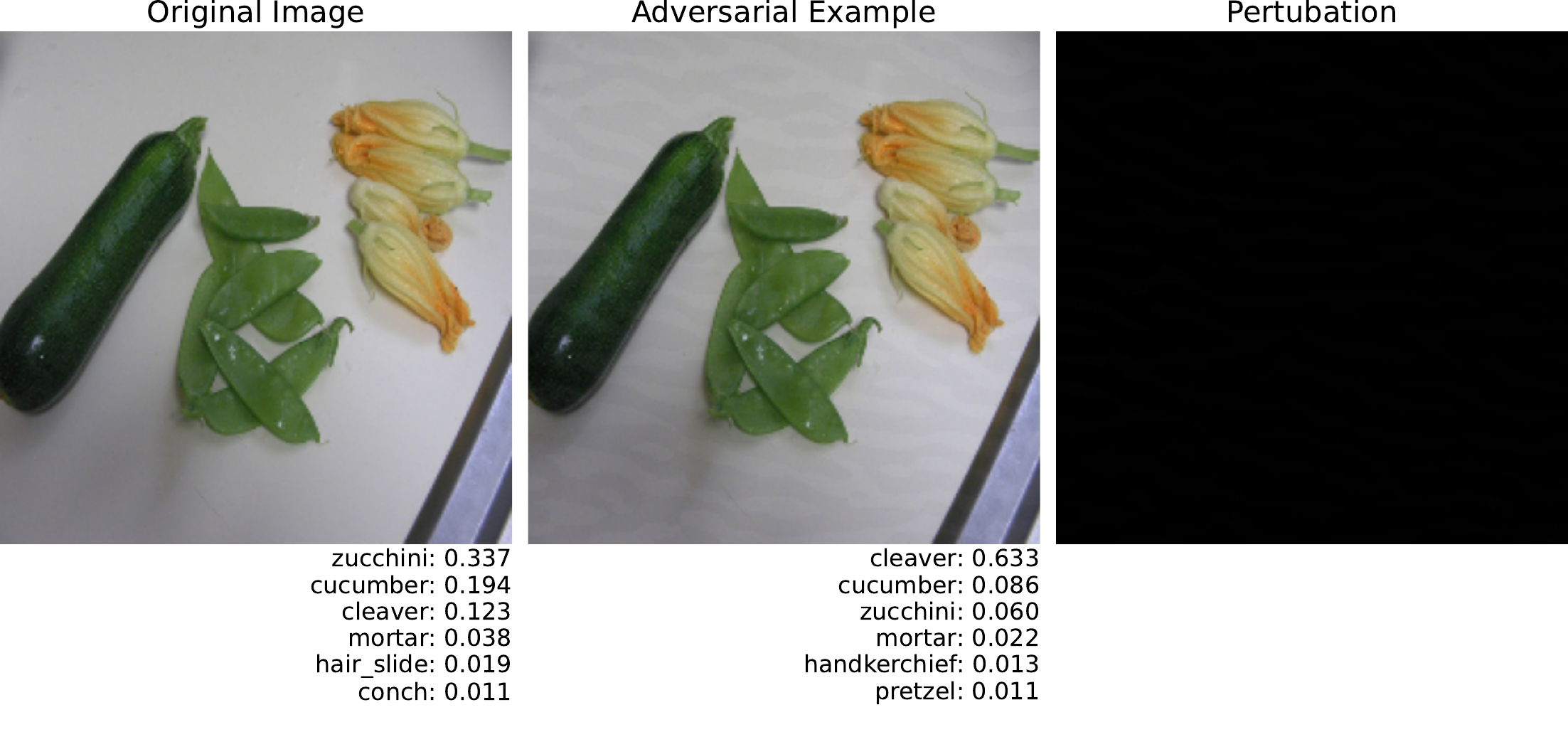}
     \caption{BO-DBA with Gabor generator}
    \end{subfigure}
\end{figure}
\begin{figure}\ContinuedFloat
     \centering
    \begin{subfigure}{0.9\textwidth}
     \includegraphics[width=\linewidth]{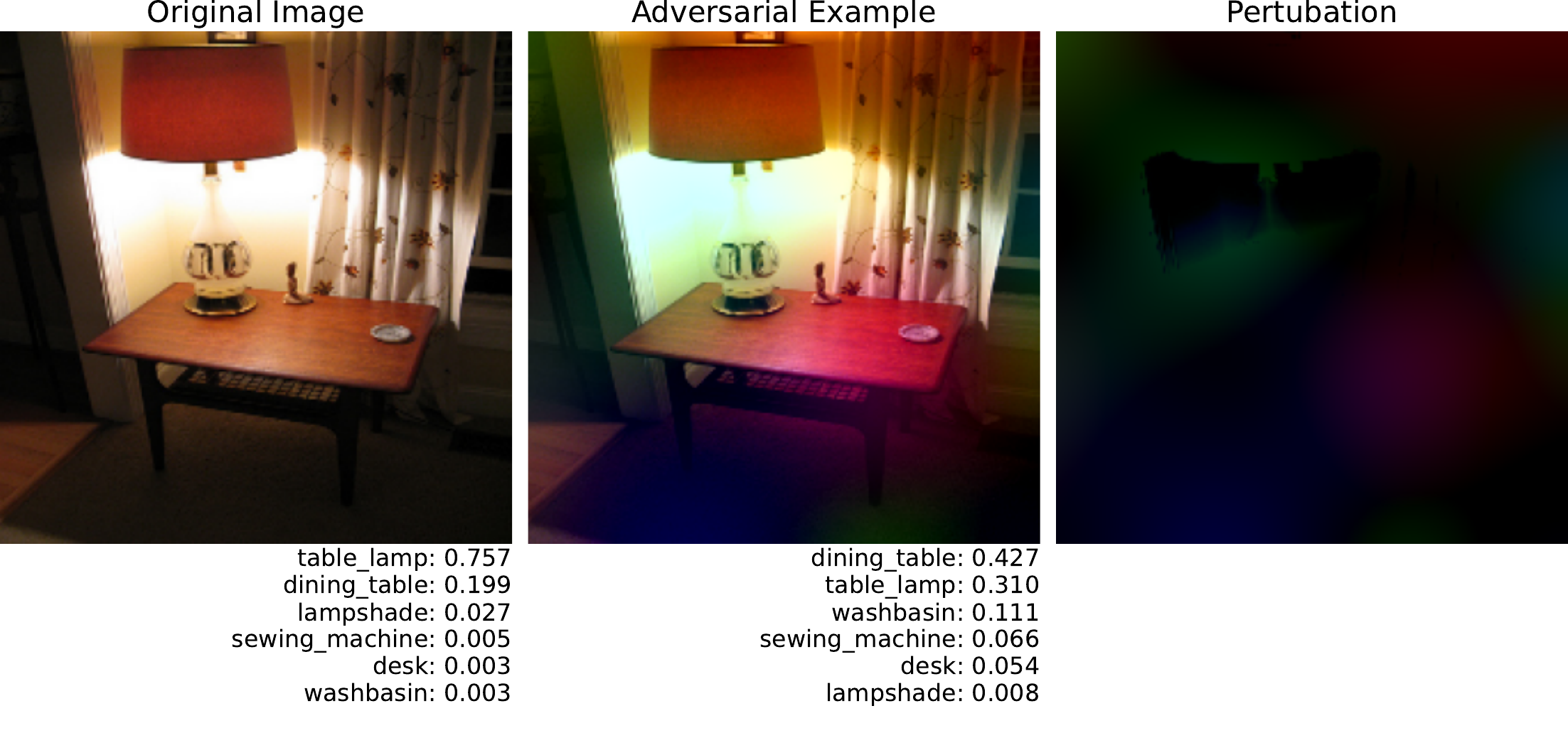}
     \caption{BO-DBA with BICU Generator}
    \end{subfigure}
\end{figure}
\begin{figure}\ContinuedFloat
    \centering
    \begin{subfigure}{0.9\textwidth}
     \includegraphics[width=\linewidth]{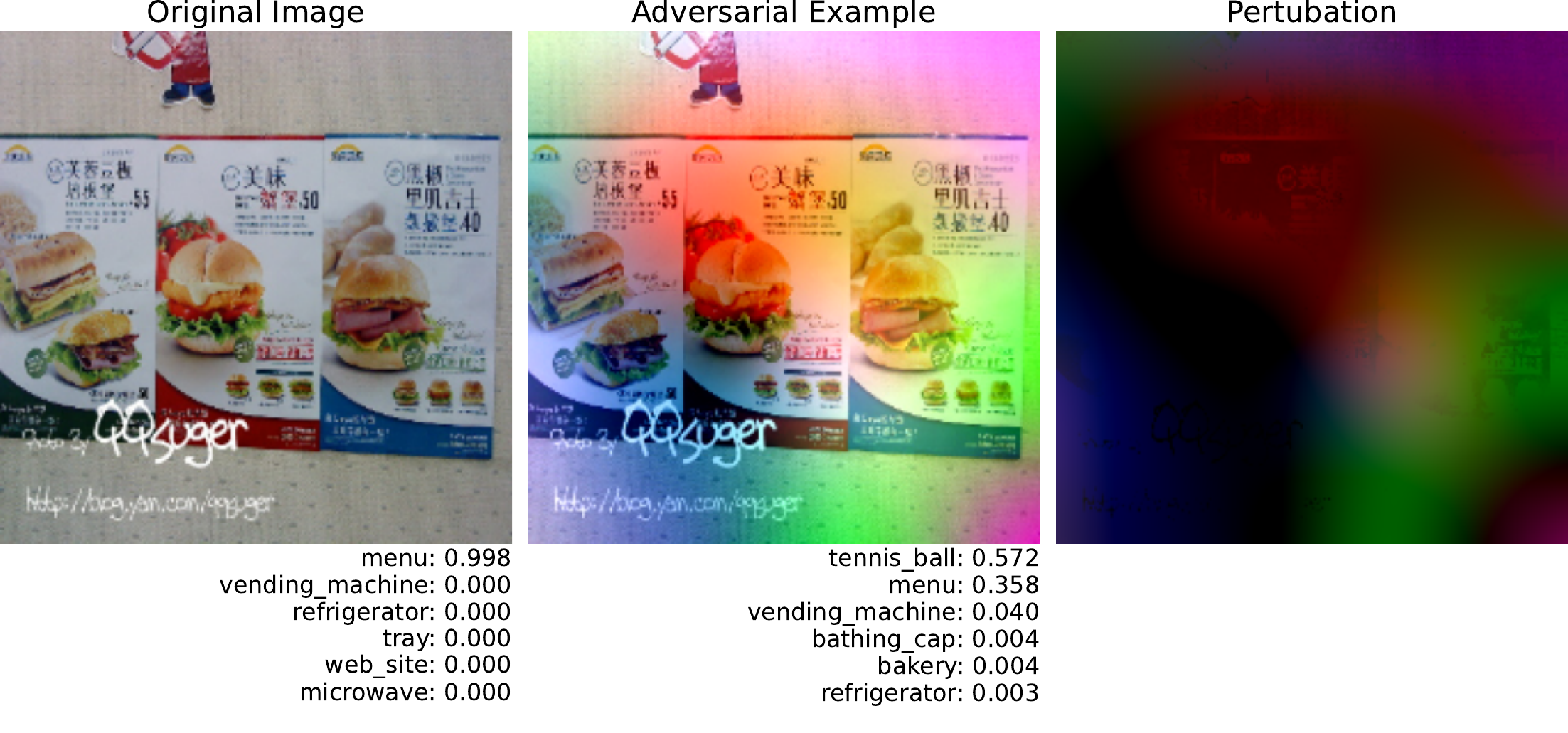}
     \caption{BO-DBA with BILI generator}
    \end{subfigure}
\end{figure}
\begin{figure}\ContinuedFloat
    \centering
    \begin{subfigure}{0.9\textwidth}
     \includegraphics[width=\linewidth]{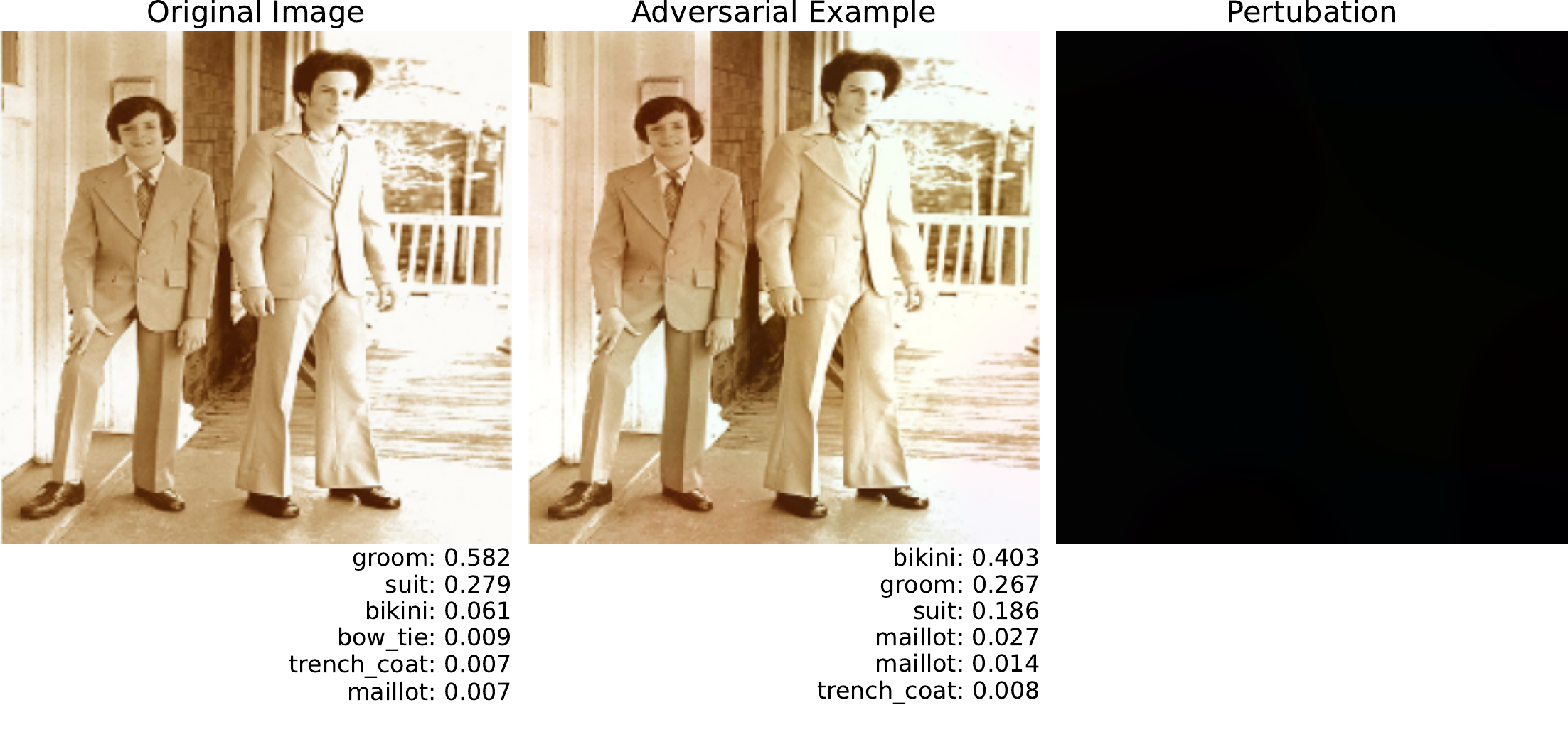}
     \caption{BO-DBA with Clustering generator}
    \end{subfigure}
\end{figure}
\begin{figure}\ContinuedFloat
    \centering
    \begin{subfigure}{0.9\textwidth}
     \includegraphics[width=\linewidth]{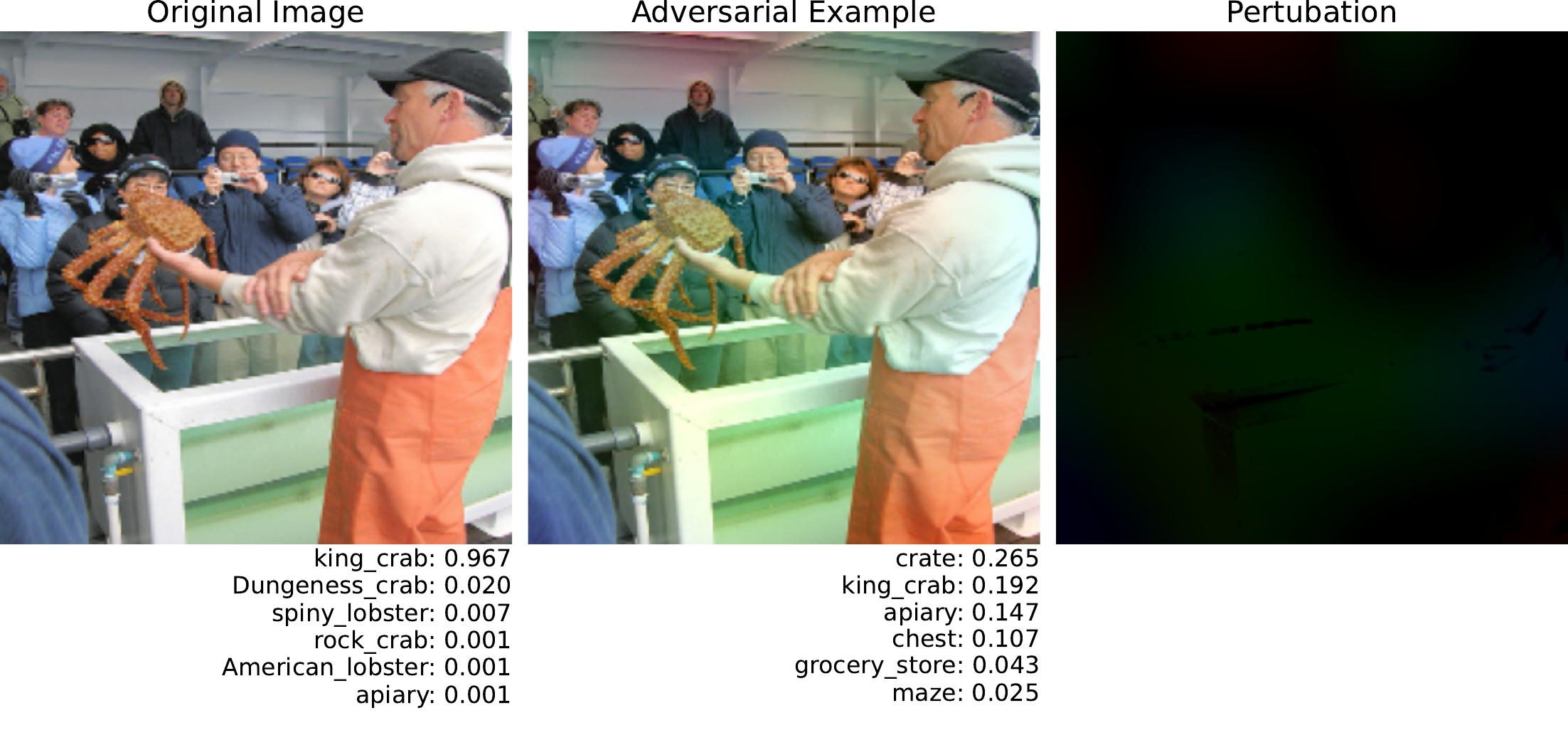}
     \caption{BO-DBA with Nearest Neighbor generator}
    \end{subfigure}
    \caption{Adversarial examples generated by BO-DBA with different pertubation generators}\label{fig:Apdx2}
\end{figure}

\end{document}